\pdfoutput=1
\RequirePackage{fix-cm}
\documentclass[smallextended]{svjour3}       
\smartqed  
\usepackage{graphicx}
\usepackage{xcolor}
\usepackage{tabularx}
\usepackage{amsmath}
\usepackage{hyperref}
\begin{document}

\title{Implementation of a neural network for non-linearities estimation in a tail-sitter aircraft}



\author{Alejandro Flores         \and
        Gerardo Flores 
}


\institute{A. Flores \at
              Centro de Investigaciones en Óptica A. C., León, Guanajuato, Mexico\\
              \email{alejandrofl@cio.mx}           
           \and
           G. Flores \at
              Centro de Investigaciones en Óptica A. C., León, Guanajuato, Mexico\\
              \email{gflores@cio.mx}  
}

\date{Received: date / Accepted: date}

\maketitle

\begin{abstract}
The control of a tail-sitter aircraft is a challenging task, especially during transition maneuver where the lift and drag forces are highly nonlinear. In this work, we implement a Neural Network (NN) capable of estimate such nonlinearities. Once they are estimated, one can propose a control scheme where these forces can correctly feed-forwarded. Our implementation of the NN has been programmed in C++ on the PX4 Autopilot an open-source autopilot for drones. To ensure that this implementation does not considerably affect the autopilot's performance, the coded NN must be of a light computational load. With the aim to test our approach, we have carried out a series of realistic simulations in the Software in The Loop (SITL) using the PX4 Autopilot. These experiments demonstrate that the implemented NN can be used to estimate the tail-sitter aerodynamic forces, and can be used to improve the control algorithms during all the flight phases of the tail-sitter aircraft: hover, cruise flight, and transition.
\end{abstract}

\section{Introduction}\label{intro}
Nowadays there exist different types of hybrid UAV's classified according to the flight mode transition principle; one of them is the tail-sitter. This type of UAV basically consists of a fixed-wing structure where the different flight modes determine the orientation of the aircraft. The transition between flight modes relies on the total rotation of the aircraft, which completely changes the dynamics of the aircraft. In general, one can say that convertible UAV's control is a current area of research since current and further applications this, relatively new, aircraft can perform like inspection of static objects in a big area. Tough the investigation of controls for convertible UAVs has been some success during the last years, it remains to tackle this problem from a practical point of view since the vast majority of the works are focused only on the theoretical part \cite{9183353}\cite{9045681}\cite{8619303}\cite{8027849}\cite{8027900}. However, such investigation has demonstrated that one of the key points for controlling convertible aircraft is the lack of knowledge of the aerodynamic forces, especially during transition phases.

Several research documents take this subject of tail-sitter control from different perspectives, which includes defining a special control algorithm for each flight mode, to develop a unified controller that works in all the flight phases (level, hover, and transition). Also, different control approaches are implemented like the common PID controller, model-free controllers \cite{doi:10.2514/6.2020-2075}; adaptive control methods also have been applied \cite{doi:10.2514/1.G003090} where input saturation and external disturbances are taken into consideration to develop a robust control algorithm. In \cite{doi:10.2514/1.G003201} \cite{doi:10.2514/1.G004520} a dynamic inversion control approach is presented where the authors consider the presence of the aerodynamics of the wing and the rotor blades to improve the performance of their control. On the other hand, the use of neural networks in the area of UAVs has its own research works. The most relevant are: \cite{8676108} and \cite{doi:10.2514/1.C034232} in which uses different types of NN to perform a dynamic inversion in a multi-rotor UAV. In \cite{9099826} a neural-networks-based control scheme is proposed to perform a flight mode transition of a ducted fan VTOL; in such a case the NN is used to compensate disturbances and uncertainties in the system.

\begin{figure}
    \centering
    \includegraphics[width = 0.8\columnwidth]{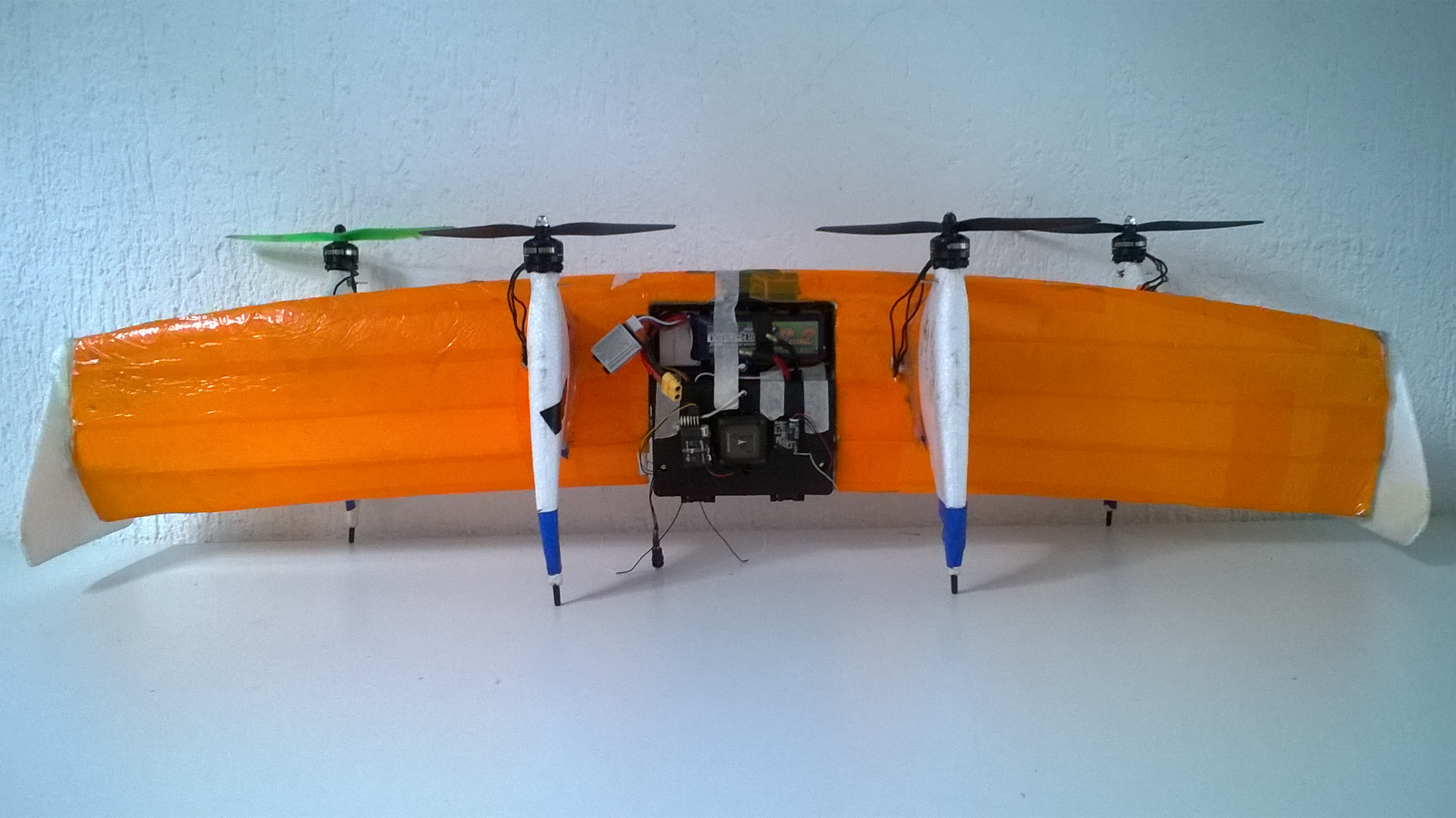}
    \caption{Tail-sitter model in which the neural network will be implemented.}
    \label{fig:tail_sitter}
\end{figure}

The objective of this work is to implement and test the performance of a simple perceptron multi-layer neural network running on the up-to-date PX4 firmware. This NN is trained to estimate the aerodynamic forces generated by the tail-sitter aircraft depicted in Fig. \ref{fig:tail_sitter} according to the body velocities during the flight tests. The final purpose of this estimation is the feedback linearization to enhance the effectiveness and simplicity of the flight control algorithm. Our approach is tested in a software-in-the-loop realistic simulation, where an optimized NN is programmed into the PX4 autopilot without the need for a companion computer.

The remainder of this short paper is as follows: in sec. \ref{sec:problem} is described the problem this work focuses and how it is planned to be solved, next in sec. \ref{sec:Results} is detailed the system model and the parameter wwe want to estimate with the neural network, also it is presented the construction and training of this network. Then, the results obtained by the SITL simulations applying the NN are presented in sec. \ref{sec:Results}, finally the conclusions are described in the sec. \ref{sec:Conclusion}.

\section{Problem setting}
\label{sec:problem}
\begin{figure}
    \centering
    \includegraphics[width = 0.6\columnwidth]{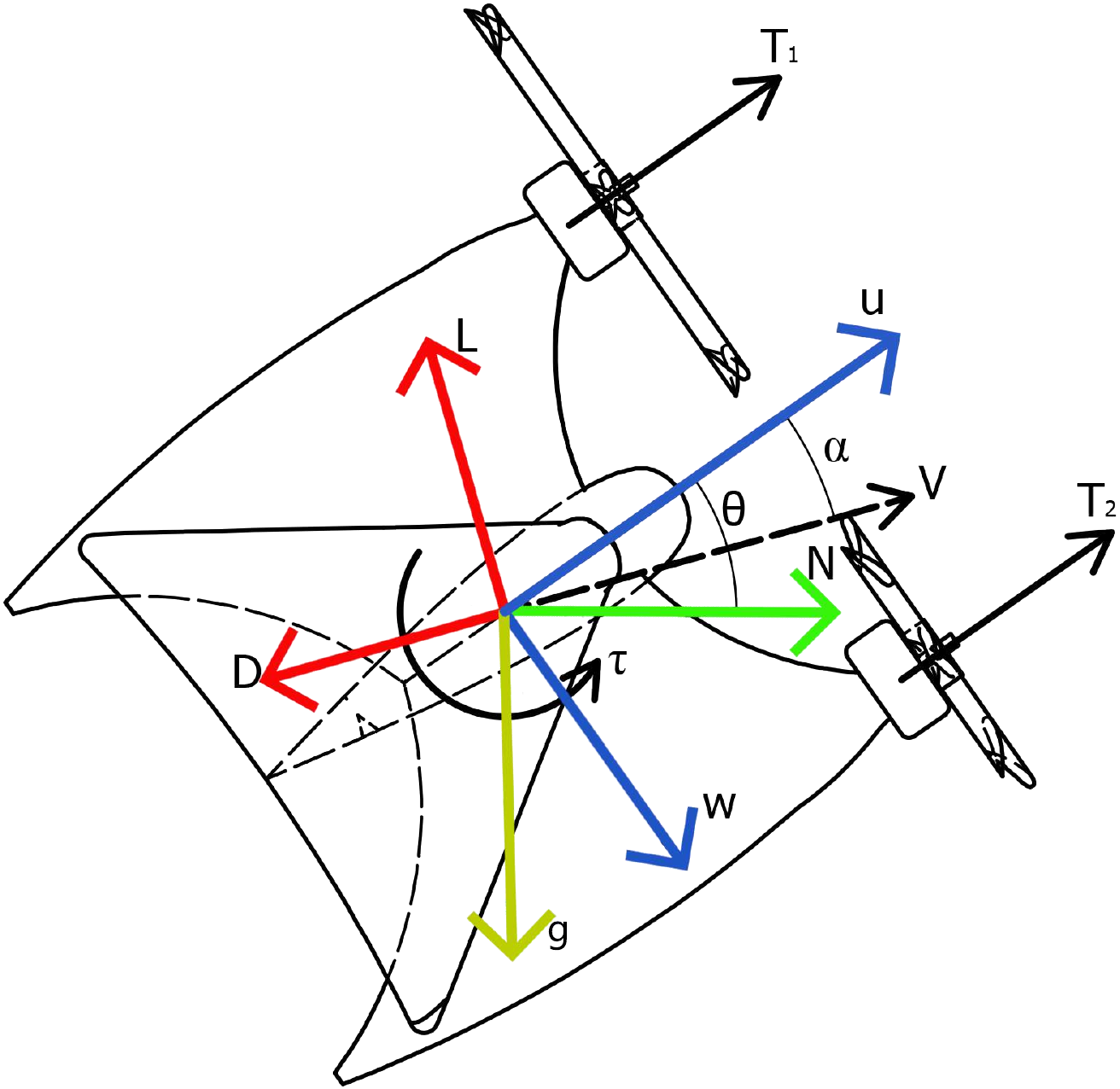}
    \caption{Free-body diagram of the longitudinal model of the tail-sitter aircraft, where the forces, moments, states and inputs are depicted.}
    \label{fig:forces}
\end{figure}

In our previous work \cite{802700}, we investigate the development of a recurrent neural network (RNN) to obtain the required state function for a feedback linearization of the dynamics of the system. In that work, we take into consideration the fact that an RNN has the capability to predict future system's behavior. Now in this paper, we take a different approach, in particular, we construct and implement a multi-layer perceptron NN in which, according to the body velocities $u,w$ (please see Fig \ref{fig:forces}) it is possible to estimate the aerodynamic forces generated by the tail-sitter.

\subsection{System model}
Nowadays the study of tail-sitter UAVs has several points of view: modeling, state estimation, control, and aerodynamic characteristics, but in most cases there is a common factor on them: the aerodynamic influence on the tail-sitter dynamics. Regarding the mathematical representation of tail-sitters, some works put attention in the model using airspeed and the path angle as states, as it is common in fixed-wing aircraft modeling. Since our case of study is a tail-sitter aircraft, we claim that a better-appropriated model is using as states the body velocities. To start with, we propose a longitudinal model representing the tail-sitter's dynamics focused on the dynamic behavior in the $x-z$ plane, that is, the model system will be resented with the horizontal and vertical velocities as described next
\begin{align}
	\Sigma : &
	\begin{cases}
	\label{eq:model_pos} 
		\, \dot{u} = \frac{1}{m} \left( T -D \cos\alpha + L\sin\alpha \right) - g\sin\theta - qw \\
		\, \dot{w} = \frac{1}{m}\left( -D \sin\alpha - L\cos\alpha \right) + g\cos\theta + qu
	\end{cases}\\
	\Omega : &
	\begin{cases}
	\label{eq:pitch}
		\, \dot{\theta} = q \\
		\, \dot{q} = \frac{1}{J} \tau
	\end{cases}
\end{align}
where $u$ and $w$ are respectively the vertical and horizontal aircraft velocities expressed in the body frame; $D$ and $L$ are the drag and lift aerodynamic forces, respectively; $\theta$ is the pitch angle, and $q$ is the rate; $\alpha$ represents the UAV's angle of attack (AoA); $m$ and $J$ represent the UAV mass and its inertia in the $y$-axis, respectively; and $\tau$, $T$ are the pitching moment and thrust, both considered as control inputs. In our work, we are mainly interested in the translational dynamics of the tail-sitter (system $\Sigma$). For that, we assume that the attitude system $\Omega$ is stable by the application of an appropriated control $\tau$; such a controller can be found in our previous work \cite{802700}.

\subsection{Problem statement}
The estimation and knowledge of the forces generated by the aerodynamics of a wing in an tail-sitter UAV have an important role in the control of the aircraft during all their flight phases. This way we propose to use a neural network capable of estimate this forces to possible be used for the control algorithm.

\section{Main result}
\label{sec:Results}
\begin{figure}
    \centering
    \includegraphics[width = \columnwidth]{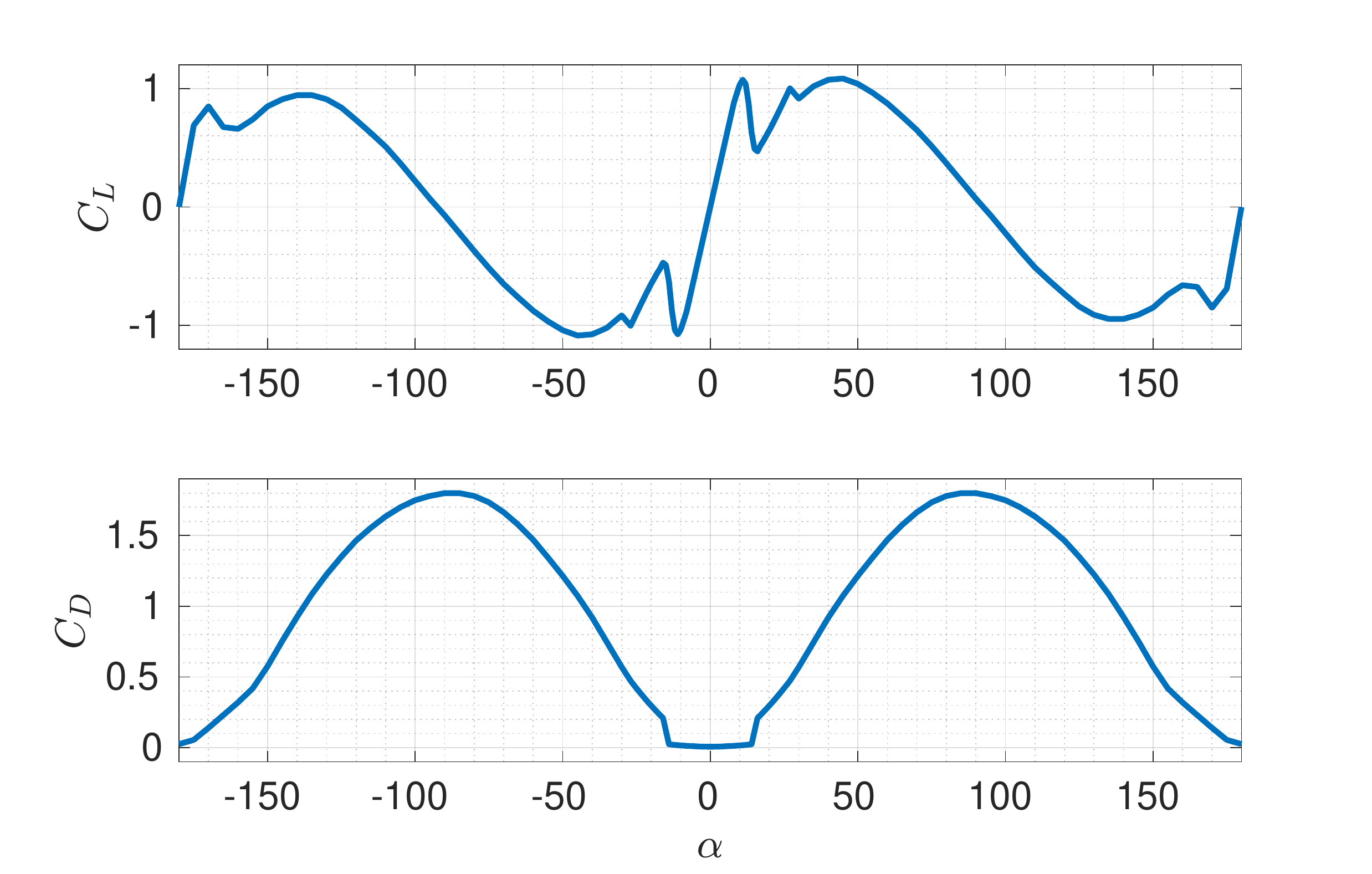}
    \caption{Lift and Drag coefficients of the airfoil NACA-0012 in the 360 degrees range of angle of attack. Due to that this airfoil has a symmetric shape, its coefficients also show symmetric patterns.}
    \label{fig:coeficientes}
\end{figure}

Notice that in system $\Sigma$ the aerodynamic forces $L$ (lift) and $D$ (drag) are given by
\begin{equation}
\label{eq:lift_drag}
L = \frac{1}{2}C_{L} V^{2} \rho S ,  \ \textrm{and } \ D = \frac{1}{2}C_{D} V^{2} \rho S
\end{equation}
take a fundamental role in the tail-sitter's accelerations. In \eqref{eq:lift_drag} $V$ is the airspeed, $\rho$ is the air density and $S$ is the wing surface area; the lift and drag coefficients $C_L$ and $C_D$ are related to the airfoil, in our case we use a NACA-0012 with coefficients depicted in Fig \ref{fig:coeficientes}.
Then, rewriting system $\Sigma$ into the form
\begin{eqnarray}
\label{eq:udotm}
    \dot{u} &=& f_{1} + g\cos{\theta} + T\\
    \dot{w} &=& f_{2} + g\sin{\theta},
\end{eqnarray}
where $f_1$ and $f_2$ contain all the forces acting in their respective body axis, i.e. $f_1 = -D \cos\alpha + L\sin\alpha - qw$ and $f_2 = -D \sin\alpha - L\cos\alpha + qu$. Since the function $f_1$ and $f_2$ mostly depends on $u$ and $w$, there is a possibility to use a NN that can estimate this functions by only using the $u$ and $w$ velocities has the neural network inputs.

\subsection{Neural network configuration}
\begin{figure}
    \centering
    \includegraphics[width = 0.8\columnwidth]{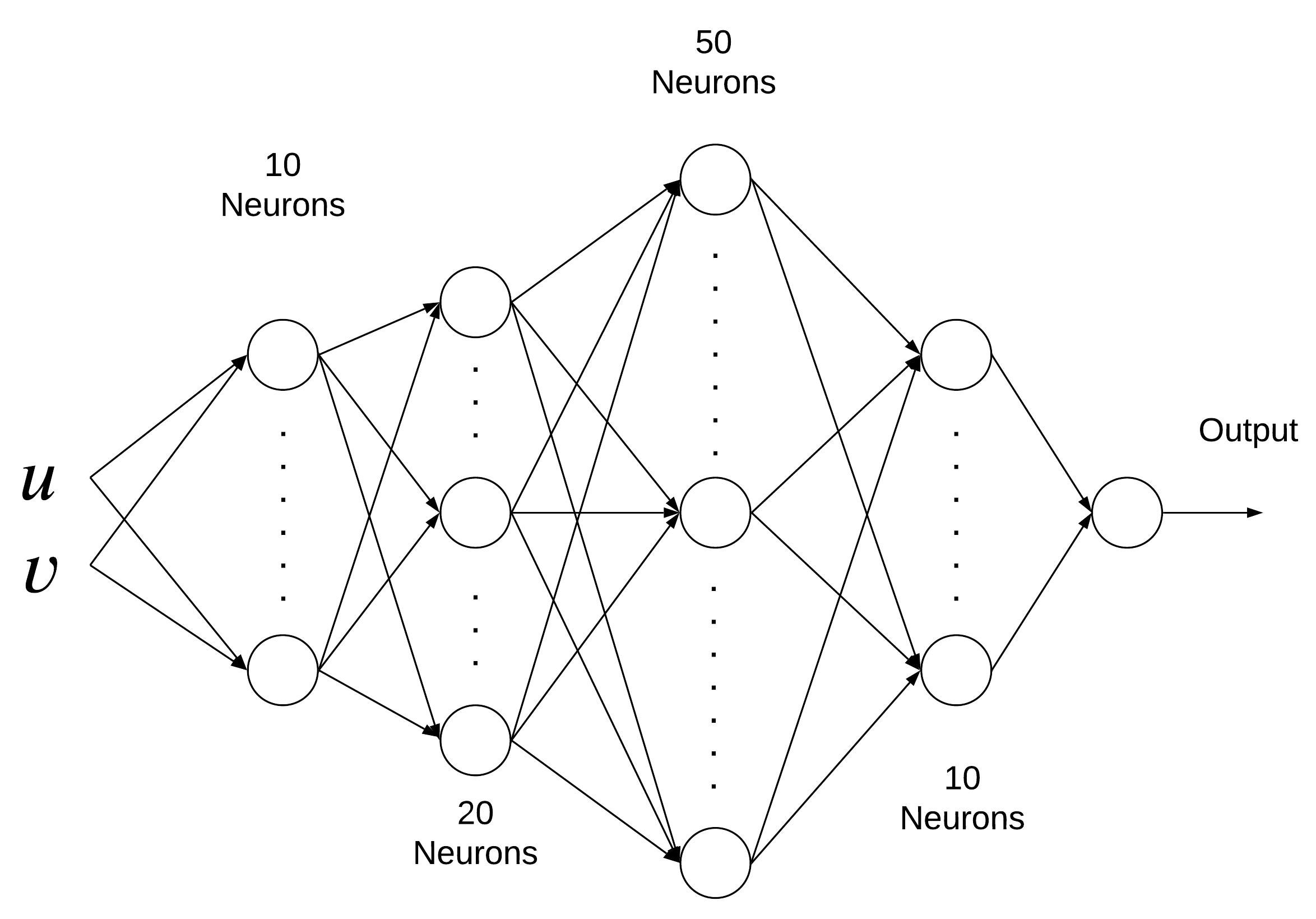}
    \caption{Neural network structure used to estimate the functions $f_1$ and $f_2$ in system $\Sigma$ represneting the body velocities of the tail-sitter aircraft. This NN consist in 5 layers with (10,20,50,10,1) neurons each one respectively. The output layer is of just one sole layer.}
    \label{fig:NN_structure}
\end{figure}

We aim to estimate the nonlinearities $f_1 (u,w)$ and $f_2 (u,w)$ in \eqref{eq:udotm} using a proposed NN configuration presented next. The structure of the implemented neural network in the PX4 firmware consists of 4 hidden layers with 10, 20, 50, and 10 neurons on each layer, respectively. The output layer consists of one layer; in Fig. \ref{fig:NN_structure} it is depicted the proposed structure for the NN. 
To choose the best activation functions it is important to have in mind that the estimation we want to obtain is in $\mathrm{R}$. We highlight that this problem is not of the kind of classification; to achieve an estimate of $f_1$ and $f_2$ we configure two different activation functions: the first one is a bounded function $f(x) = tanh(x)$ where the output values are within $(-1,1)$. Such an activation function is applied in the second and third layer, and in the rest layers it is applied a linear activation function $f(x) = x$. Tab. \ref{tab:NN_structure} shows the principal characteristics of the neural network. 
\begin{table}[ht]
	\centering
	\caption{Neural Network Structure.}
	\label{tab:NN_structure}
	\noindent
    \begin{tabularx}{0.9\textwidth}{X X X X}
			\hline\hline\\
			[-3mm]
			Layer & Neurons & N Weights & Activation function\\
			\hline
			{1} & {10} & {2} & {$f(x) = x$}\\
			{2}	& {20} & {200} & {$f(x) =tanh(x)$}\\
            {3}	& {50} & {1000} & {$f(x) =tanh(x)$} \\
            {4} & {10} & {500} & {$f(x) = x$}\\
            {5} & {1} & {10} & {$f(x) = x$} \\
            [1mm]
			\hline\hline
		\end{tabularx}
\end{table}

\subsection{Neural network training}
As it is well known, to train any neural network it is required a dataset of input values and its corresponding output results. To obtain our dataset we perform flight simulations of a modeled tail-sitter, where random inputs $(u,v)$ were applied to simulate and get the outputs $(f_1,f_2)$, doing this, we obtained 10,000 data samples to train the NN.

The used training method was the back-propagation algorithm, in which the weights update take an action from the last neural network layer to the first one, according to the learning rate and error obtained in the output. The training process consisted of 10 epochs until the weight does not show any change. The total number of weights that were calculated for this neural network is 1,712 float values, which means that all these values must be stored in the PX4 Autopilot to be used during the control execution.
\begin{figure}
    \centering
    \includegraphics[width = 0.7\columnwidth]{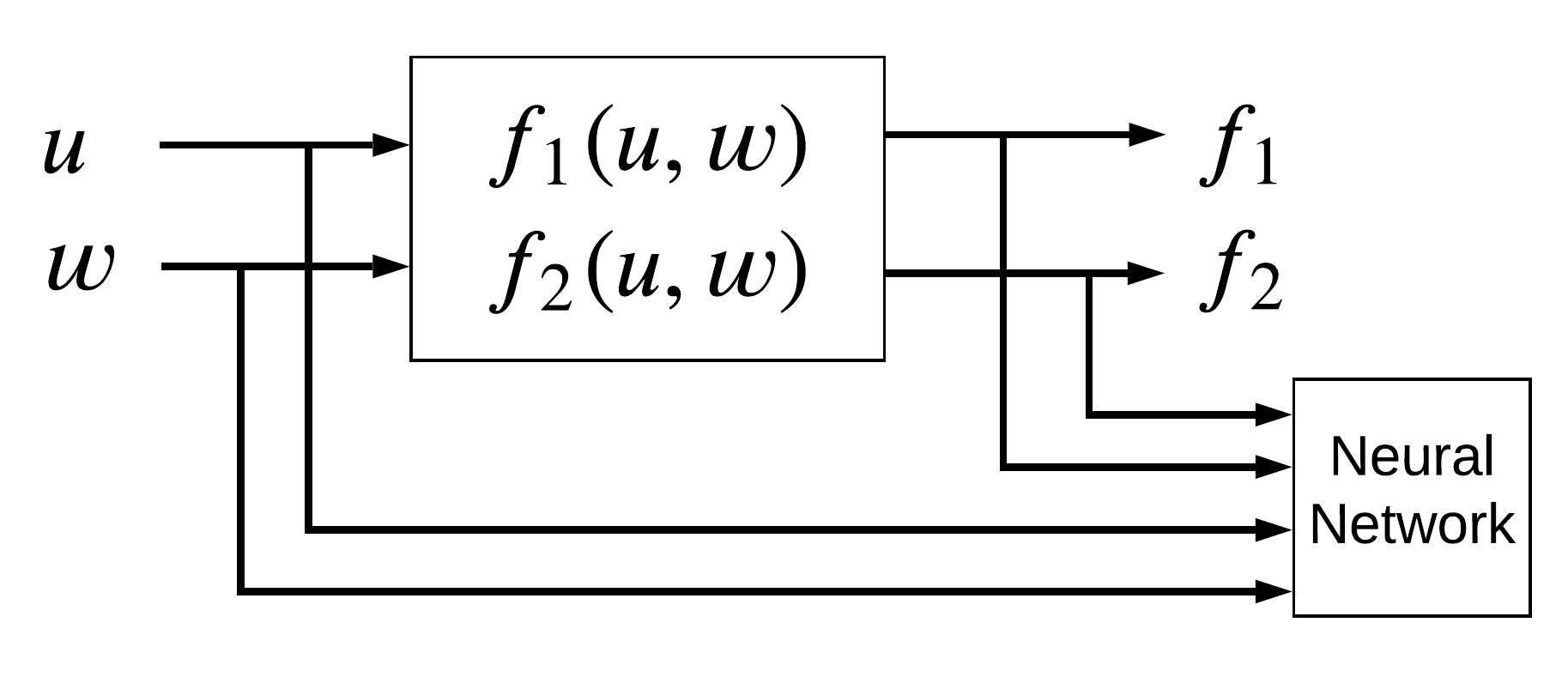}
    \caption{Block diagram that represents the training dataset for the implemented neural network. Since¿ we use a supervised neural network training, the dataset should include a collection of all the possible kinds of inputs $(u,w)$ and their respective outputs $(f_1,f_2)$ that will help to the neural network to achieve the weight's optimal value.}
    \label{fig:my_label}
\end{figure}
\begin{figure}
    \centering
    \includegraphics[width = 0.9\columnwidth]{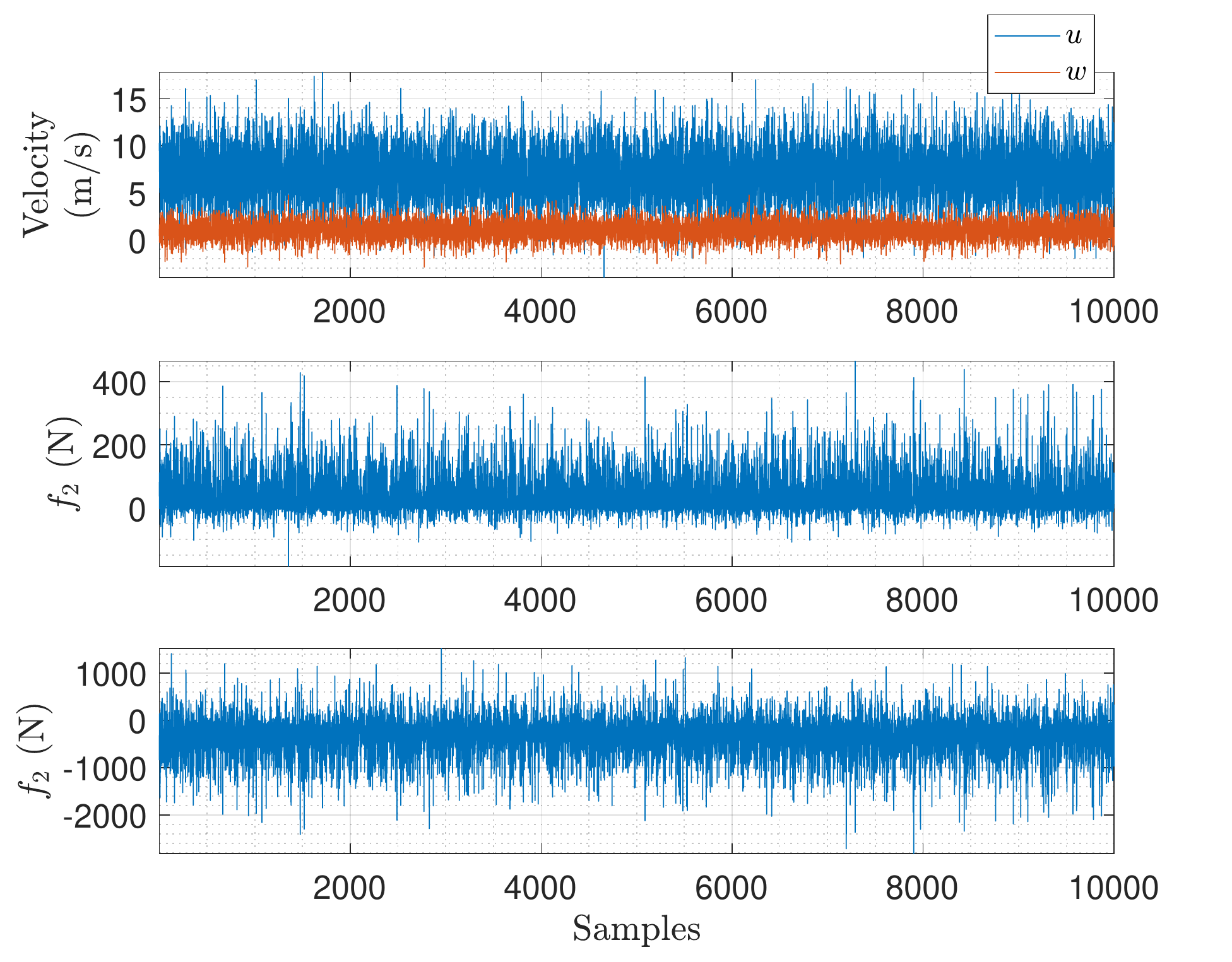}
    \caption{Input and output data-set generated by simulation, as we can see, we generate a random bounded data input $(u,w)$ according to the nominal velocities that a tail-sitter UAV performs, ín this case the velocities goes, in $u$ from 0 m/s to 18 m/s where arte the hover ans level flight modes respectively. In teh case of the velocity $w$, this values are from -1 m/s to 4 m/s which corresponds to slow displacements during hover flight mode.}
    \label{fig:my_label}
\end{figure}

\section{Results}
\label{sec:Simulation}
To verify the real performance of our approach, we conduct several simulations in the Gazebo software \cite{1389727} taking the available tail-sitter UAV model and run a software-in-the-loop (SITL) simulation. This simulation consists in to emulate the physics and aerodynamics in a realistic CAD model. In addition, the computer emulates the micro-controller running the PX4 firmware with the Autopilot to finally make a virtual representation of the real system. It is important to mention that, this type of simulation executes the real controller that will be run in real experiments, the functionality and results are very close to real behavior.

The simulation consists of a flight test running the neural network during all the flight time. We perform a flight experiment consisting of autonomous take-off, followed by hovering, then transition, followed by cruise flight, and finally returning to hover flight for achieving a landing; all these are conducted autonomously. During all the flight envelope the proposed neural network estimate the functions $f_1$ and $f_2$ in the system $\Sigma$.
\begin{figure}
    \centering
    \includegraphics[width = 0.8\columnwidth]{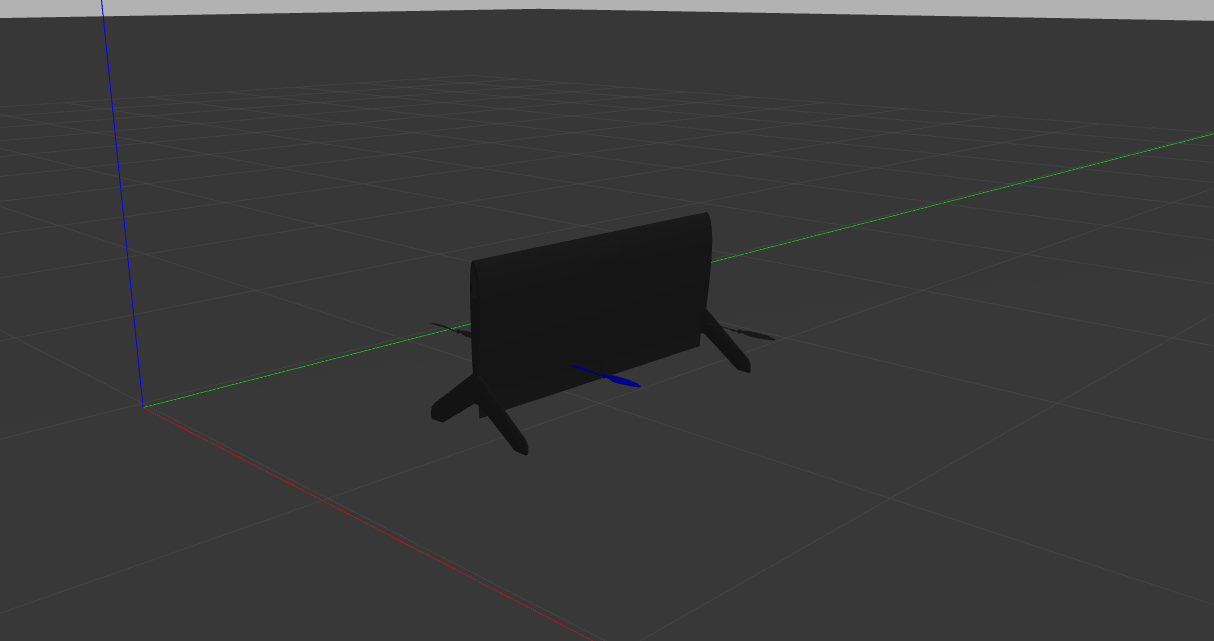}
    \caption{Software in the loop (SITL) simulation in the Gazebo environment using the tail-sitter UAV model for the implementation of the neural network. The experiment board all the flight modes performed autonomously in this order: take-off, hover, transition, cruise, transition, hover, and landing.}
    \label{fig:hover_sim}
\end{figure}

\begin{figure}
    \centering
    \includegraphics[width = \columnwidth]{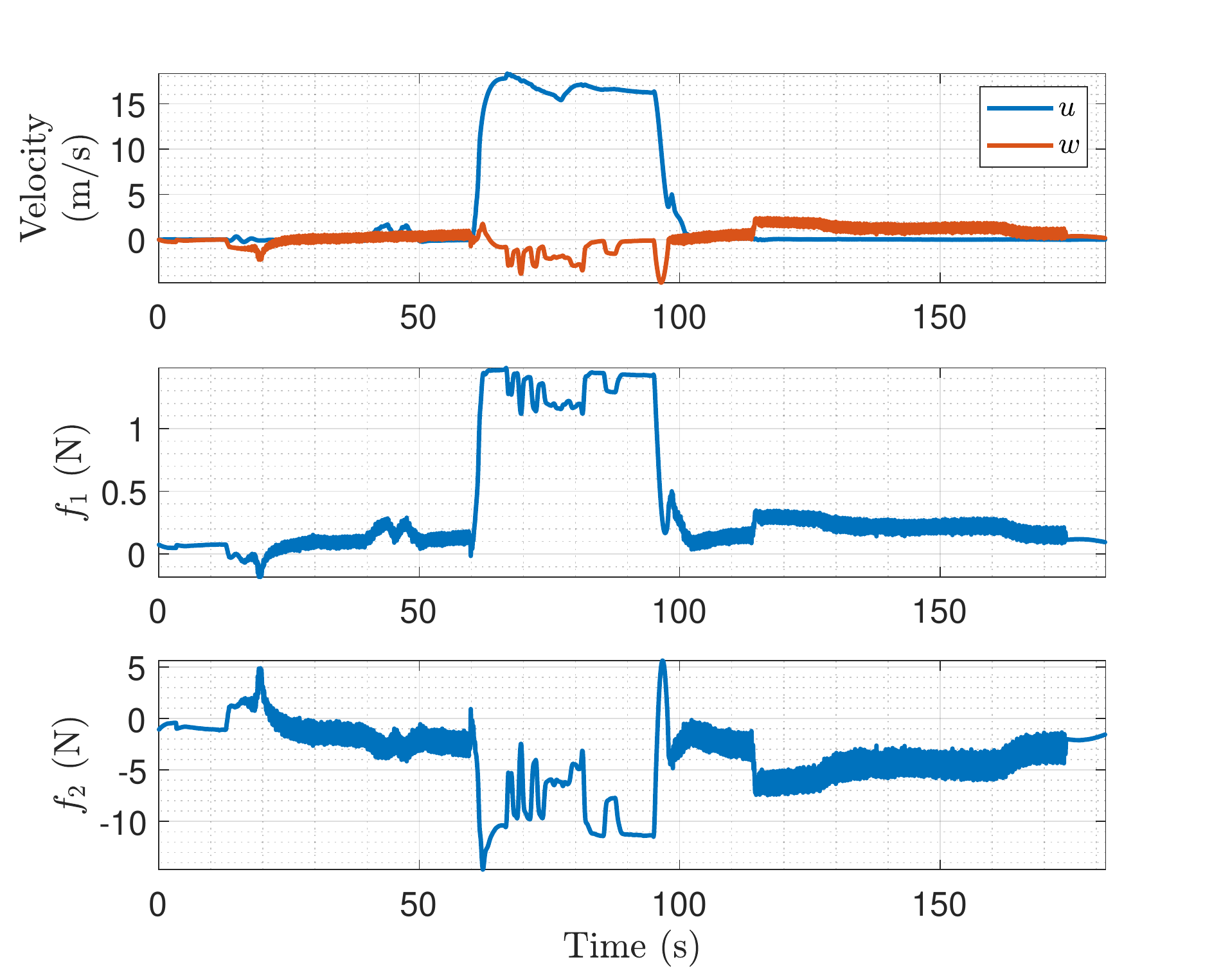}
    \caption{Forces $f_1$ and $f_2$ estimated during a tail-sitter flight. In this simulation, the two flight modes were performed to see the behavior of the forces generated during both flight modes.}
    \label{fig:f1f2_estimation}
\end{figure}

The results of the simulation are shown in Fig. \ref{fig:f1f2_estimation} where the upper graph depicts the actual tail-sitter velocities $(u,w)$. The middle and the lower graph show the neural network estimated aerodynamic forces $(f_1,f_2)$, respectively. The beginning of the test consists of the vertical take-off and a short hover flight mode period (from 0s to 60s), then the velocities $u$ and $w$ corresponds to small values as depicted in the upper graph. By consequence, the aerodynamic forces generated by such velocities must correspond to small values as described in the two last graphs of Fig. \ref{fig:f1f2_estimation}. The second part of the flight test consists in the level flight mode, this part takes place from 60s to 100s. During the cruise flight mode, the velocity $u$ must be higher than in hover; this is due to the UAV's attitude and the high-speed requirements in cruise flight. Conversely, $w$ must keep small values compared to $u$. In the first graph of Fig. \ref{fig:f1f2_estimation}, when $t=60$s, $u$ significantly increases while $w$ does slightly. Because of this, it is logical to expect that the aerodynamic forces $f_1$ and $f_2$ increases. This can be seen in the next two graphs in the same figure, where $f_1$ that interacts in the direction of $u$ depicts an increased value; in the case of $f_2$ this value is higher since it compensates the gravity force and keeps the UAV in the air. Finally, the last flight stage consists of a return to hover flight performing some small displacements as it is shown in Fig. \ref{fig:f1f2_estimation}. The small horizontal movements during hovering cause forces in $f_2$ due to the friction of the wing with the air when it is perpendicular to the displacement. The modified code for the NN implementation can be accessed \href{https://github.com/LAPyR/NN_aerodynamic_estimation}{here}. 

\section{Conclusion}
\label{sec:Conclusion}
In this work, a neural network was implemented in the actual PX4 firmware developed by PX4 autopilot. This NN is responsible to estimate non-linear terms due to aerodynamics variation in the tail-sitter aircraft. Once the estimation of these elements it was possible to compensate them to finally simplify the flight control algorithm. The obtained results show that the proposed NN estimates the non-linearities of the system. The restriction of using a micro-controller difficult a possible implementation of a complex NN to improve the estimation of non-linear dynamics due to the low memory capacity of such devices and also because of the processing velocity required to execute the whole flight code.

From the given experiments, we claim that using artificial intelligence methods together with simple control algorithms composed of feed-forward terms can control systems with high variability in its dynamics, as is the case of a tail-sitter UAV. |


%
%



%
%


\end{document}